%% file: main.tex
\pdfoutput=1

\documentclass[conference]{IEEEtran}

\usepackage{cite}
\usepackage{graphicx}
\usepackage{xspace}
\usepackage{url}
\usepackage{mathtools}

\usepackage[table]{xcolor}
\usepackage{microtype}
\usepackage{array}
\usepackage{tabularx}
\usepackage{booktabs}
\usepackage{multirow}
\usepackage{float}

\usepackage{sansmath}
\usepackage{tikz}
\usepackage{pgfplots}

\usepackage{amsmath,amssymb}
\usepackage[ruled,vlined]{algorithm2e}

\pgfplotsset{compat=1.13}
\usepgfplotslibrary{fillbetween}
\usepackage[flushleft]{threeparttable}

\newcolumntype{Y}{>{\centering\arraybackslash}X}
\newcommand\YUGE{\fontsize{24}{30}\selectfont}
\newcommand{\car}{\textit{Cardea}\xspace}
\newcommand{\mimic}{\textsc{Mimic-iii}\xspace}
\newcommand{\kaggle}{\textsc{Kaggle}\xspace}
\newcommand{\framew}{framework\xspace}

\begin{document}

\title{Cardea: An Open Automated Machine Learning Framework for Electronic Health Records}

\author{\IEEEauthorblockN{
Sarah Alnegheimish\IEEEauthorrefmark{1}\IEEEauthorrefmark{2},
Najat Alrashed\IEEEauthorrefmark{1},
Faisal Aleissa\IEEEauthorrefmark{1}, 
Shahad Althobaiti\IEEEauthorrefmark{1},
Dongyu Liu\IEEEauthorrefmark{2},
Mansour Alsaleh\IEEEauthorrefmark{1} \\ and
Kalyan Veeramachaneni\IEEEauthorrefmark{2}}
\IEEEauthorblockA{\IEEEauthorrefmark{1}The Center for Complex Engineering Systems at KACST and MIT, Riyadh, Saudi Arabia.}
\IEEEauthorblockA{\IEEEauthorrefmark{2}MIT, Cambridge MA, USA. \\
\{smish, dongyu, kalyanv\}@mit.edu, \{nalrashed, fsaleissa, salthobaiti, maalsaleh\}@kacst.edu.sa}
}

\maketitle

\begin{abstract}
   An estimated 180 papers focusing on deep learning and EHR were published between 2010 and 2018. Despite the common workflow structure appearing in these publications, no trusted and verified software framework exists, forcing researchers to arduously repeat previous work. In this paper, we propose \car, an extensible open-source automated machine learning framework encapsulating common prediction problems in the health domain and allows users to build predictive models with their own data. This system relies on two components: Fast Healthcare Interoperability Resources (FHIR) -- a standardized data structure for electronic health systems -- and several \textsc{AutoML} frameworks for automated feature engineering, model selection, and tuning. We augment these components with an adaptive data assembler and comprehensive data- and model-auditing capabilities. We demonstrate our framework via 5 prediction tasks on \mimic and \kaggle datasets, which highlight \car's human competitiveness, flexibility in problem definition, extensive feature generation capability, adaptable automatic data assembler, and its usability.
\end{abstract}


\IEEEpeerreviewmaketitle

\input{manuscript/introduction.tex}
\input{manuscript/background.tex}
\input{manuscript/cardea.tex}
\input{manuscript/exp.tex}
\input{manuscript/results.tex}
\input{manuscript/userstudy.tex}
\input{manuscript/discussion.tex}
\input{manuscript/conclusion.tex}

\section*{Acknowledgment}
This work was supported by the Center for Complex Engineering Systems at King Abdulaziz City for Science and Technology (KACST) and Massachusetts Institute of Technology. This work was also partially supported by the National Science Foundation Grants ACI-1761812.

\bibliographystyle{IEEEtran}
\bibliography{references}

\clearpage

\appendix
\input{manuscript/supplementary_material.tex}

\end{document}

%% file: manuscript/introduction.tex
\section{Introduction}

Over the last decade, the number of data-driven machine learning models developed to tackle problems in healthcare has increased tremendously~\cite{xiao2018opportunities,steyerberg2008clinical}.
Numerous studies have demonstrated machine learning's effectiveness for predicting readmission rates, emergency room wait times, and the probability of a patient not showing up for an appointment \cite{Futoma_2015,Goffman2017ModelingPN,liu_length_2010}.

However, a hospital or entity trying to build a model must start from scratch -- a difficult process given that, at present, there is a significant shortage of data scientists with the requisite machine learning skills. 
Moreover, starting from scratch means retreading existing work: repeating essentially the same steps as previous studies, facing similar limitations and pitfalls, and re-engineering strategies to overcome them. 
Even worse, hospitals may complete this process only to find that gaps in data prevent their newly-built model from achieving the accuracy promised by the studies. 

To address this complicated and troubling situation, we ask: ``\textit{Could there be a simple way to share software that enables development of healthcare models more broadly?}'' Imagine this scenario: A user decides that s/he needs a model for predicting readmissions. S/he has access to a data engineer with programming and data management experience but not necessarily machine learning skills. 
The two of them access a community-driven, verified software platform to help develop a machine learning model from raw data. After two or three simple functional calls, they build a model and obtain numerous metrics, reports and recommendations. Such a workflow is the goal of our framework, which we call \car.

The integration of many sources of electronic health records via the wide scale adoption of the Fast Healthcare Interoperability Resources (FHIR) standardized format (schema) was intended to enable the development of reusable machine learning-based models~\cite{khalilia2015clinical,rajkomar2018scalable,hoffman2018intelligent,kasthurirathne2015towards}. But while reusability was the original goal of the FHIR schema, the inevitable variance in the amount and quality of available data requires a framework that can adapt to missing tables and data items. In addition, the framework should be able to run the full prediction process using only available information. 

Additionally, the past few years have seen the rise of automated approaches to machine learning model development. However, the construction of a powerful framework requires automation to work end-to-end: defining the initial prediction task, creating the features required for machine learning, and facilitating machine learning model development and tuning. Thus \car needed to define abstractions and intermediate data representations that allow the software to not be bound by intricacies specific to a particular data at hand.  

Moreover, the system must be able to adapt to varying data availabilities and context-specific intricacies. For example, two hospitals with different subsets of tables and fields may spur an \textsc{AutoML} method to choose entirely different features and pipelines without manual intervention. The method must also allow users to tweak and extend predefined problems by adding functionality and/or modifying corresponding parameters. It should also enable advanced users to easily define and contribute new prediction problems, and allow the community to evaluate the contribution's adherence to the abstractions.

In this paper, we present \car, an open-source framework for automatically creating predictive models from electronic healthcare data. \textbf{Our contributions in this paper are}:
\begin{itemize}
    \item[--] The first ever open-source, end-to-end automated machine learning framework for electronic healthcare data.~\footnote{Our software is available at \url{https://github.com/DAI-Lab/Cardea}}.
    \item[--]A set of key abstractions and data representations designed by carefully scrutinizing hundreds of machine learning studies on healthcare data. These abstractions allow us to transport data from its raw form to a predictive model while capturing metadata and statistics. 
    \item[--] An end-to-end framework incorporating an adaptive data assembler, an automated feature extractor, and an automated model tuning system. 
    \item[--] Through numerous case studies, we show the ability of the framework to adapt to different scenarios across multiple healthcare datasets. These datasets include \mimic -- an accessible, openly available and multi-purpose dataset -- and the \kaggle dataset.
    \item[--] Through case studies, we also present the framework's competitiveness when compared with humans performance across many dimensions: model accuracy, reduction of software complexity, and ability to support numerous tasks humans can envision for the data. 
\end{itemize}

The rest of this paper is organized as follows: We describe related work in the field in Section~\ref{sec:back}, followed by the description of \car and its components in Section~\ref{sec:cardea}. We demonstrate some use cases and their results in Sections~\ref{sec:exp-settings} \&~\ref{sec:results}. We report our user study in Section~\ref{sec:userstudy}. Finally, Section~\ref{sec:discussion} presents discussion and Section~\ref{sec:conclusion} concludes the paper.

%% file: manuscript/background.tex
\section{Background and Related Work}

\label{sec:back}
\subsection{FHIR} 
There is a vast amount of literature proposing various standards for the electronic exchange of health information, most taking interoperability and integration as main objectives. 
Health Level 7 (HL7) introduced several of these standards in 1989; their first contribution is still used in over 90\% of US hospitals~\cite{lamprinakos2014}.

Most recently, HL7 introduced FHIR, a new standard that aims to transcend the shortcomings of previous standards~\cite{fhir}. 
FHIR has been adopted across many technologies in the health industry, including mobile applications, prediction software, and health management systems. 
FHIR has been used to communicate patient information in addition to reporting prediction results~\cite{khalilia2015clinical}; for scalable development of deep learning models~\cite{rajkomar2018scalable}; and to provide streamlined access to data for the development of mortality decision support applications~\cite{hoffman2018intelligent}. 
Finally, FHIR standards are being integrated with several health management systems~\cite{kasthurirathne2015towards}.

\subsection{Predictive Health Automation}
The continuous collection of various patient data (demographics, lab tests, visits, procedures, drugs administered, etc.)~\cite{xiao2018opportunities} has driven an increase in the number of models and algorithms tackling health-related problems. Xiao et al. identified over 180 publications related to deep learning and EHR between 2010 and 2018.

The health-predictive models include: patient health prediction~\cite{moons2012risk,hunt2018multi,tiwari2017automatic}; health-sector operations~\cite{riascospredicting}; predicting patient mortality based on the competing risks that patients may experience over time~\cite{nori2015simultaneous}; and predicting the progression of Alzheimer's in a patient~\cite{hunt2018multi}. In~\cite{tiwari2017automatic}, the authors classify the severity of a radiology report based on the report's textual features.  The authors of~\cite{riascospredicting} utilize insurance claim data to predict a patient's length of stay. 

In~\cite{chakraborty2019robustibm}, the authors built a framework to identify survival risk factors. 
Their work demonstrates how automatic feature generation can surpass expert factors in performance.
In addition, they highlight the impact of data-driven approaches in which they were able to deploy a machine learning model in $< 1$ month, in comparison to the estimated 1-3 year development period.
Computational Healthcare is another endeavor developed to analyze health data, generate statistics, and train and evaluate machine learning models~\cite{chlib}.
However, although the tool is open-source and available online, it has not been updated over the past three years.
Moreover, both frameworks support a limited number of prediction problems and are not generalizable across different data. They are instead tailored around a specific machine learning library, which limits their extensiblity. 

%% file: manuscript/cardea.tex
\section{Cardea System}

\label{sec:cardea}

\begin{figure*}[ht]
    \centering
    \includegraphics[scale=1.05]{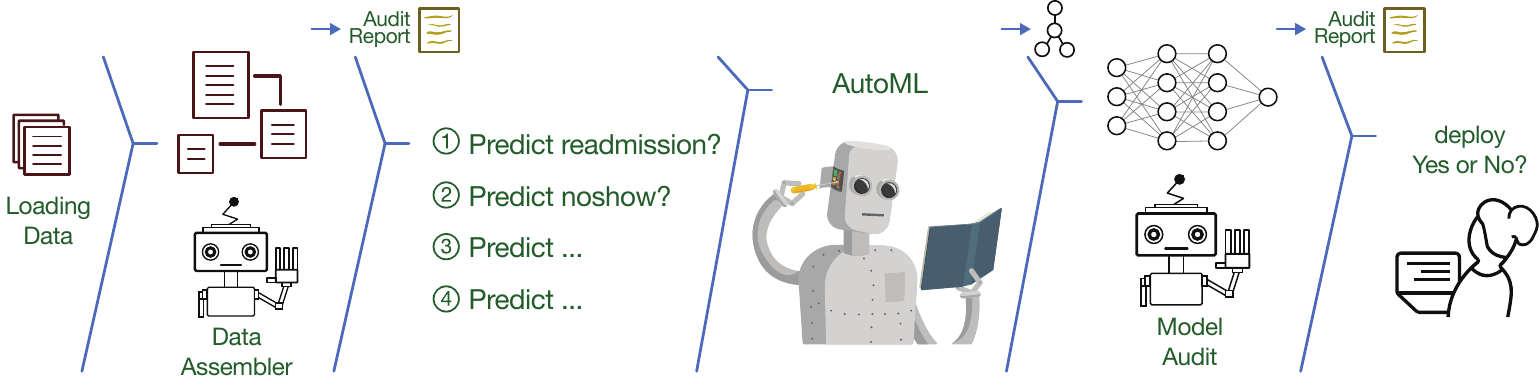}
    \caption{Cardea: A fully automated open-source framework for learning predictive models from electronic healthcare records. Cardea consists of multiple automated components. A automatic data assembler loads and organizes the data. A data-audit module automatically assesses the quality of different parts of the data. A set of prediction problems is stored in the library, allowing users to choose from them and set some of their parameters. An \textsc{AutoML} component generates and tunes training examples, features, labels, and models. A model-audit module appraises the resulting model along various metrics.}
    \label{aipm1}
    \vspace{-6pt}
\end{figure*}

\car is an automated framework for building machine learning models from electronic health care records. Users first load their dataset into the framework through an adaptive data assembler. The data auditor generates a report that summarizes any discrepancies within the data. The user then selects a prediction problem to tackle from a list of predefined problems. Next, the framework starts the \textsc{AutoML} phase, which consists of three components: feature engineering, model selection, and hyperparameter tuning. Finally, the framework helps the user to audit the model's results by reporting the model's performance. 
Figure~\ref{aipm1} depicts the framework's workflow.
\footnote{ We describe the data and the model auditor in the appendix ~\ref{auditing}.} 

\begin{figure*}[ht]
    \centering
    \includegraphics[scale=1.2]{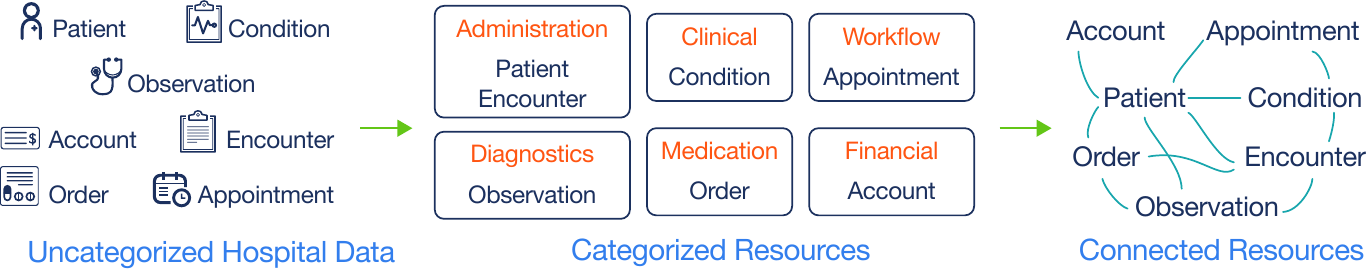}
    \caption{The FHIR standard can represent any health data in a categorized container format, which simplifies the exchange of data between connected resources. For example, $S^{p}$ (\texttt{patient}) and $S^{e}$ (\texttt{encounter}) fall under the administration module. 
    $S^{c}$ (\texttt{condition}), which captures the health conditions for a patient, falls under the clinical module and is linked to the $S^{p}$.}
    \label{fhir_rep}
    \vspace{-6pt}
\end{figure*}

\subsection{Adaptive Data Assembler}
A typical healthcare dataset captures information about the various day-to-day aspects of a hospital, including administration, clinical records, drugs, and financial information. The extensive nature of this data creates a complex structure of intertwined information that could lead to disparate data organization schemas. 
FHIR is an international standard that helps to organize the exchange of hospital data~\cite{fhir}. 
Resources (tables), which are the basic building blocks in FHIR, define all exchangeable content for a particular resource, which then falls under a module, as shown in figure~\ref{fhir_rep}. 
In addition, possible connections with neighboring resources are preset in each resource's metadata.
The metadata identifying these relationships, table names, field names, and their types are available at: \url{https://hl7.org/fhir/}. 

Notationally, for a given dataset, we have resources, $\mathcal{S}$, $S^{1..N}$, where each resource $S^{k}$ comprises variables $v_{1..j}$. We denote a variable $v_{j} \in S^{k}$ as $S^{k}_{j}$. To determine a relationship between $S^{1}$ and $S^{2}$ we form a dependence notation $S^{1}_{a}\rightarrow S^{2}_{b}$ which illustrates that $S^{1}$ holds a foreign key $v_{a}$ that references the primary key $v_{b}$ in $S^{2}$. This can be further simplified to $(S^{1}_{a}, S^{2}_{b}) \in \mathcal{R}$, where $\mathcal{R}$ is the set of all possible relations over $\mathcal{S}$.
The structure is analogous to a directed graph $\mathcal{G} = \langle \mathcal{S}, \mathcal{R} \rangle$ with vertices $\mathcal{S}$, and edges $\mathcal{R}$. 
Therefore, any data with an underlying relational database structure can be expressed in the framework as graph $\mathcal{G}$.

Although most standard schemata are comprehensive -- for example, FHIR has a total of 492 resources with 2342 relational links -- in almost all the scenarios we faced, we found that only a subset of the overall data was available to each hospital. Thus what data is available varies significantly. 
% For example, the data we worked with for Section~\ref{sec:results} covered merely $\sim 0.5$\% of resources.
Therefore, as mentioned in Algorithm~\ref{algo:dataloader}, an adaptive data assembler must accept any data $\mathcal{D}$ where $\mathcal{D} \subseteq \mathcal{S}$ and load $D^{k}_{a}$ into $S^{k}_{a}$ if $v_{a}\in D^{k}$ otherwise it continues to load the remaining $v_{b..j} \in D^{k}$; excluding nonexistent variables. 
Next, the algorithm tests loaded variables and adds possible relationships $\{(S^{k}_{i}, S^{l}_{j})\; | \; \forall \;l \le k\}$ s.t. $(S^{k}_{i}, S^{l}_{j}) \in \mathcal{R}$. In our current implementation, when a user uploads the data, the data assembler creates an in-memory representation (dictionary) of the metadata for only the resources loaded and relevant foreign keys, which we call an \texttt{entityset} $E$\footnote{\texttt{EntitySets} is a in memory representation of multiple connected tables introduced by the open source tool - \texttt{Featuretools}. \url{https://docs.featuretools.com/}}.
In each iteration, the algorithm checks for a cycle graph in the relationships. If a conflict arises, the loader calls the function $\psi$($\cdot$, $\cdot$) to alter the structure of $E$ and $R$. The method receives any operation in which it consolidates the root resource with another and breaks the causing link in the least intrusive approach.
Such a procedure is critical for adapting a metadata structure that conforms to the original schema, while resolving the challenges of foreign keys with missing primary keys.

% Alternitively, we can think of entityset $E$ as hypergraph $\mathcal{H}$ with vertices $S$ and edges $R$, $\mathcal{H} = (\mathit{S}, \mathit{R})$.

\begin{algorithm}[ht]
\SetAlgoLined
  \KwData{set of ingested data $\mathcal{D}$}
  \KwResult{entityset $E$ of organized data, and relationships $R$}
  
  \For{$S^{k} \in \mathcal{S}$}{
    \For{$v_{i} \in S^{k}$}{
      \uIf{$v_{i} \in D^{k}$}{
        $S^{k} \gets S^{k} \cup D^{k}_{i}$\\
        $R \gets R \cup \{(S^{k}_{i}, S^{l}_{j}) \;|\; \forall \;l \le k , (S^{k}_{i}, S^{l}_{j}) \in \mathcal{R}\}$
      }
      \uElse{
        % $\forall v_{j} \in R $
        % $\text{\textbf{drop }} S^{k}_{i}$\;
        $S^{k} \gets S^{k} \cup \{\phi\}$
      }
    }
    % \uIf{\texttt{check\_cycle($R$)}}{
    $E(k) \gets S^{k}$\\
    \uIf{$\exists \:(S^1, \dots, S^k) \in R \wedge S^1 = S^k$}{
        $[E(k), R] \gets \psi$($E$, $\bowtie$)
    }
  }
  \KwRet{$E + R$}\;
  \caption{Adaptive data assembler} \label{algo:dataloader}
\end{algorithm}

\subsection{Specifying Prediction Problems}
\label{sec: pred_prob}
\begin{figure}[ht]
    \centering
    \includegraphics[width=0.8\linewidth]{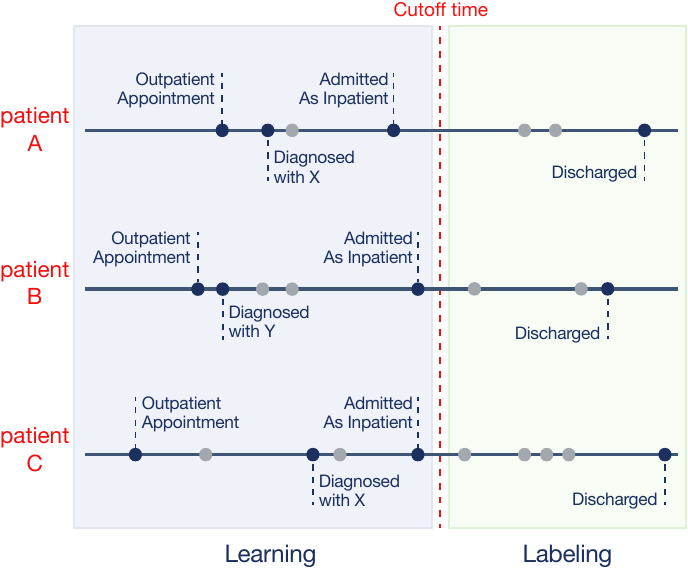}
    \caption{An example of cutoff time. Any events before a certain time are used for learning, while events after that time are used for labeling.}
    \label{cutoff}
\end{figure}

For most machine learning models, users define a specific  outcome they would like to predict. They then prepare the data by identifying and generating labeled training examples along with time points for each training example. These time points separate the data that can be used to learn the model from the data that can't be used. Authors in \cite{kanter2016label} note that this process underpins most prediction tasks across several domains; thus, formulating a prediction task and creating training examples can be further abstracted. Here is a breakdown of how \car undertakes these three steps:

\noindent \textbf{Generating cutoff times}: The cutoff time is a timestamp that splits the data into two segments: before and after. The data before the cutoff time is used to learn the model, while the data after is used for labeling. This is important because it ensures that predictive models are trained on data that does not already contain label information or other future information that is not usually available at the time of prediction, a problem widely known as label leakage. Cutoff times can be generated in several ways. In some cases, they are set based on the problem definition. For example, the cutoff time in a readmission problem is usually the time of discharge for each patient, whereas the cutoff time for a length of stay problem would be the time of admission. In \car, the method \texttt{gct} extracts pre-specified cutoff times, but the user also has the ability to overwrite the algorithm and create custom cutoff times. Figure~\ref{cutoff} shows an example of a uniform cutoff time across several patients. 
        
\noindent \textbf{Writing a labeling function}: Given a relational dataset, the cutoff time, and the target entity for which the prediction is sought, the labeling function $f(\cdot)$ is written to produce a label or target value. 
    
\noindent \textbf{Creating labeled training examples}: Given a list of cutoff times for multiple entities, this method iterates over the cutoff times and generates labels for each entity-instance in the list.

After these three steps are complete, we have $\mathcal{X}$, what we call ``\texttt{label\_times}'' - a tuple of $<e_{id}$, $t_c$, $l>$. With these \texttt{label\_times}, the task of defining a prediction problem in \car is generalized.  The problem definition is configured once, and it can be later stored and reused in the framework by any user. The procedure of these steps is detailed in Algorithm~\ref{algo:probdef}. For a given entityset $E$, and a target entity $E(k)$, the target label or outcome that one wants to predict may already exist as a column called target label $v_{l}$. Otherwise, a labeling function enables computation of the label for each entity instance. For every $e_{id} \in E(k)$ we specify a cutoff time $t_{c}$ as the start time or end time of an event + an offset (e.g. 24 hours) using \texttt{gct}, depending on the nature of the problem. To avoid label leakage, any event that occurs after that time will not be used to generate features.

% As described in Algorithm~\ref{algo:probdef}, for a given entityset $E$, target entity $E(k)$ and target label $v_{l}$, we can generate the cutoff times for any specific prediction problem. For every $v_{l} \in E(k)$ we specify a cutoff time $t_{c}$ as the start time ($t_{s}$) or end time ($t_{e}$) of an event, depending on the nature of the problem. For example, for the appointment no-show problem, the cutoff time is specified as the time the appointment was made. To avoid label leakage, any event that occurs after that time will not be used to generate  features. In addition, if $v_{l}$ is not present in the $E(k)$, we can then update $E$ (denoted by $E_{u}$) and calculate $v_{l}$ using a labeling function $f(\cdot)$ and recursively generate the cutoff times.

\begin{algorithm}[!ht]
\SetKw{Init}{Initialize}{}
\SetAlgoLined
  \KwData{entityset $E$, target $k$, target label $v_{l}$}
  \KwResult{set of tuples $\mathcal{X}$}
  \Init{$\mathcal{X} \gets \{\phi\}$, offset $\epsilon$}\\
    \For{$e_{id} \in E(k)$}{
        $t_{c} \gets $ \texttt{gct($e_{id}, \epsilon$)}\\
        \uIf{$v_{l} \notin E(k)$}{
            $l \gets f(E, t_c, e_{id})$\\
        }
        \uElse{
            $l \gets v_{l}(e_{id})$\\
        }
        $\mathcal{X} \gets \mathcal{X} \cup {<e_{id}, t_{c}, l>}$
    }
    \KwRet{$\mathcal{X}$}
  \caption{Generating Label Times}\label{algo:probdef}
\end{algorithm}

We present an initial list of predefined problems. These include predicting no-show appointments, length of stay, readmission, diagnosis prediction, and mortality prediction. This list will be expanded to include a broader set of prediction problems.

\subsubsection{Appointment No-Show}
\label{sec:appointment no-show}
Predicting no-show appointments can help hospitals optimize their scheduling policies \cite{Goffman2017ModelingPN}. 
This problem is defined as a binary classification problem, as it predicts whether or not a patient will attend an appointment.

\subsubsection{Length of Stay}
\label{sec:los}
The system approaches this problem two ways: as a classification problem that determines which patients stayed more than a week from the point of admission~\cite{rajkomar2018scalable}, and as a regression problem that predicts the number of days a patient is expected to stay in a hospital~\cite{liu_length_2010}.

\subsubsection{Readmission}
\label{sec:readmission}
Predicting patient readmission can help a hospital better manage its resources and measure its own performance and quality \cite{readmission_article}. This is implemented as a binary classification prediction that predicts whether a patient will revisit the hospital within 30 days from the date of discharge~\cite{doi:10.1093/intqhc/mzx011}.

\subsubsection{Diagnosis Prediction} 
Patient diagnosis prediction can help with the planning of intervention and care as well as with the optimization of resources~\cite{pols_outcome}. The user provides a diagnosis code, and \car generates a target label according to whether the patient has received the specified diagnosis since the point of admission.

\subsubsection{Mortality Prediction}
\label{sec:mortality}
Hospital performance can be measured by predicting mortality as a performance indicator~\cite{Gomes2010MortalityPM}. 
In the case where the mortality label is not present, the framework extracts this information through a list of underlying causes of death according to the visits' diagnosis code, which includes motor vehicle accidents, intentional self-harm (suicide) and assault (homicide), then generates a target label based on this list of codes. 
\car treats this as a classification problem where it predicts patient mortality from the point of admission.

\subsection{Auto Machine Learning}
\label{sec: Auto Machine Learning}
Data scientists typically follow a fairly similar procedure for prediction problems. This repetitive effort motivated the automation of the machine learning workflow (e.g.,~\cite{AML1,AML2,AML3}), which includes two main phases: first, the featurization of data, and second, modeling and tuning hyperparameters. 

\subsubsection{Feature Engineering}
\car utilizes the relational nature of the entityset $E$ to perform feature synthesis~\cite{kanter2015deep}. We use automated feature engineering tool called \texttt{featuretools}. 
Given an entityset, for each entry, corresponding to the entity instance $e_{id}$ in \texttt{label\_times}, \texttt{featuretools} automatically performs two steps:
\begin{itemize}
    \item Removes the data past the corresponding $t_c$ for that entry. 
    \item Computes a rich set of features by aggregating and transforming data across all the entities. This is accomplished by executing an algorithm called Deep Feature Synthesis. For more details we refer the reader to \cite{kanter2015deep} and the Featuretools library itself available at: \url{https://github.com/FeatureLabs/featuretools}.
\end{itemize}
 
Users can control the type of features and the number of features generated by Featuretools through specifying hyperparameters - such as aggregation primitives, depth, etc. The output of the feature engineering is a matrix of each entry $e_{id}$ with its engineered features. We combine the features generated with the labels extracted in algorithm~\ref{algo:probdef} to obtain a dataset that is perfectly formatted for the succeeding component.

\subsubsection{Modeling and Hyperparameter Tuning}
Considering the copious amount of machine learning algorithm libraries (e.g., \texttt{scikit-learn}, \texttt{Tensorflow}, \texttt{Keras}, and their \textsc{AutoML} versions), we opted to give the user the flexibility and freedom of using the library of their choice. 
We are able to accommodate for this dynamic structure while maintaining an interpretable format by using primitives and computational graphs as proposed in~\cite{smith2019machine}.
The user specifies the preprocessing, postprocessing, and ML algorithm s/he is interested in. 
In addition, we use Bayesian optimization methods to tune the hyperparameters~\cite{bergstra2013hyperopt}. This method aims to find optimal hyperparameters in fewer iterations by evaluating those that appear more promising based on past results (informed search). It achieves a better performance than random search or grid search~\cite{bergstra2011algorithms,bergstra2013making,hutter2011sequential,thornton2013auto,snoek2012practical}.

%% file: manuscript/exp.tex
\section{Experimental Settings}

\label{sec:exp-settings}
In this section, we describe the characteristics of datasets loaded into \car, elaborate on the prediction tasks selected, and detail the settings used for the \textsc{AutoML} phase.

\noindent \textbf{Datasets}: To emphasize \textit{Cardea}'s adaptivity to complex datasets, variability in data availability, and disparate prediction problems, we utilize two publicly available, widely studied complex datasets. To qualify as realistic, datasets have to be multi-table, with multiple entities, and complex. with several foreign key relationships. We used two datasets. The first, Kaggle’s Medical Appointment No-Shows dataset ~\cite{noshow}, consists of 21 variables that were loaded into \car’s data assembler to generate a total of 9 resources conforming to the FHIR structure\footnote{We converted the Kaggle dataset into FHIR schema}. \kaggle contains an approximate 20:80 class division across patients that showed up to their appointments vs. not. The second, \mimic, is a rich Electronic Health Records (EHR) dataset widely used in the research community to perform various prediction tasks. Overall, the data is composed of 40 tables, storing 534 variables and 63 relationships. 
(see appendix~\ref{app:datasets} for complete description).

\noindent \textbf{Prediction problem definitions}: 
For \kaggle, we predict whether a patient will show up to an appointment. On the other hand, \mimic includes both patient- and operations-centric information. \mimic's versatiliity allows us to test \car on a number of different prediction problems, including mortality, readmission, and length of stay.

\noindent \textbf{AutoML Settings}: \textsc{AutoML} is divided into two main components: first, feature engineering; second, modeling and hyperparameter tuning. For feature engineering, we apply central tendency and distribution operations $\oplus = \{$\texttt{sum}, \texttt{standard deviation}, \texttt{max}, \texttt{min}, \texttt{skew}, \texttt{mean}, \texttt{count}, and \texttt{mode}$\}$. In addition, we utilize time-oriented functions that extract \texttt{day}, \texttt{month}, \texttt{year}, and type of \texttt{day} (weekend/weekday). Maximum depth for extracting features is set to $N = 2$. For preprocessing, we employ simple imputation and min-max scalers to normalize the data between \texttt{[0, 1]}.
For modeling, we utilize eight classification methods and seven regression methods. 
(see appendix~\ref{app:manual_features} for full results). 
Moreover, each machine learning algorithm has a set of tunable hyperparameters used for Bayesian optimization of the model~\cite{bergstra2013hyperopt}.

%% file: manuscript/results.tex
\section{Case studies}
\label{sec:results}
Our \framew proposes a number of abstractions and data representations, and a set of automated tools, aimed at enabling the wide-scale use of machine learning to work with electronic healthcare records. In addition to performing competitively compared to a human, the \framew should be adaptable, flexible across use cases, and easy to work with, and should reduce the need for extensive software re-engineering. In the following subsections, through a series of case studies, we demonstrate \car's efficacy along these lines.

\subsection{Case study: Human competitiveness}

\pgfplotsset{compat=1.11,
    /pgfplots/ybar legend/.style={
    /pgfplots/legend image code/.code={%
       \draw[##1,/tikz/.cd,yshift=-0.25em]
        (0cm,0cm) rectangle (3pt,0.8em);},
   },
}
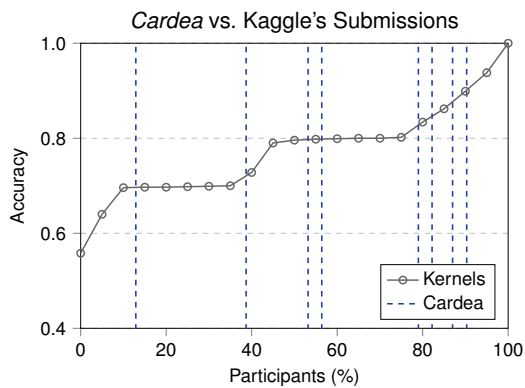
\begin{figure}[t!]
\centering

\begin{tikzpicture}[scale=0.7,font=\sffamily\sansmath] 
\begin{axis}[
width=9.7cm,height=7cm,
xlabel={Participants (\%)},
ylabel={Accuracy},
xmin=0,xmax=100,
ymin=0.4,ymax=1.0,
x label style={font=\sffamily},y label style={font=\sffamily},
y tick label style={/pgf/number format/.cd,fixed,fixed zerofill,precision=1,1000 sep={}},
ymajorgrids=true,grid style=dashed,
tick align = outside,xtick pos=left,ytick pos=left,
legend pos=south east,legend columns=1,legend cell align=left,legend style={line width=0.15mm},
 title style={at={(0.5,1.1), font=\sffamily\large},anchor=north},
 title = \car vs. Kaggle's Submissions
]
\addplot[color=darkgray!80,mark=o,solid, thick] table[x = X,y = Y, col sep=comma] {data/kaggle_updated.csv};
\addlegendentry{Kernels}
\addplot +[mark=none, dashed, draw={rgb:red,1;green,2;blue,5}, line width=0.3mm] coordinates {(12.9, 0) (12.9, 1)};
\addplot +[mark=none, dashed, draw={rgb:red,1;green,2;blue,5}, line width=0.3mm] coordinates {(38.7, 0) (38.7, 1)};
\addplot +[mark=none, dashed, draw={rgb:red,1;green,2;blue,5}, line width=0.3mm] coordinates {(53.2, 0) (53.2, 1)};
\addplot +[mark=none, dashed, draw={rgb:red,1;green,2;blue,5}, line width=0.3mm] coordinates {(56.4, 0) (56.4, 1)};
\addplot +[mark=none, dashed, draw={rgb:red,1;green,2;blue,5}, line width=0.3mm] coordinates {(79, 0) (79, 1)};
\addplot +[mark=none, dashed, draw={rgb:red,1;green,2;blue,5}, line width=0.3mm] coordinates {(82.2, 0) (82.2, 1)};
\addplot +[mark=none, dashed, draw={rgb:red,1;green,2;blue,5}, line width=0.3mm] coordinates {(87, 0) (87, 1)};
\addplot +[mark=none, dashed, draw={rgb:red,1;green,2;blue,5}, line width=0.3mm] coordinates {(90.3, 0) (90.3, 1)};
\addlegendentry{Cardea}
\end{axis}
\end{tikzpicture}
 
\caption{
Accuracy scores of \kaggle participants. The blue dotted lines indicate where \car was able to achieve an equivalent accuracy.}
\label{fig:result-barplot}
\vspace{-8pt}
\end{figure}

In this subsection, we showcase the results obtained by \car on Kaggle's Medical Appointment No-Shows dataset and compare it to 78 kernel notebooks from~\cite{noshow} with their reported accuracy for the same prediction problem.
One challenge is that individual users choose train-test splits, cross-validate, and report metrics differently. For a fair comparison with our classifiers, we report the average result over several cross-validation rounds.
Figure~\ref{fig:result-barplot} compares the framework to other users' models at each percentile. 
Overall, \car performed well compared to the human users. More specifically, the Gradient Boosting Classifier (GB), which had an accuracy of 0.91, was able to score higher than 90\% of kernel notebooks. Running the framework end-to-end automatically generates 95 features, which feed an ML pipeline that will likely outperform 80\% of existing models.

\subsection{Case study: Flexibility and coverage}
\label{sec:reproducible_pred_tasks}

One of \car's claims is that its abstractions support the formulation of any prediction problem on a healthcare dataset. To support this claim, we examine the \mimic dataset, which is popular among researchers for modeling prediction tasks in healthcare. (As of writing this paper, there are 1,278 citations of the \mimic dataset.) We surveyed a collection of 186 papers that applied machine learning to the \mimic between 2017 and 2019; of those, 90 of them had prediction tasks. 
% Hence, the rest of results will be limited within the context of those 90. 
For each of these papers, we recorded (1) the prediction task it was trying to solve, (2) the data used, and (3) the label source of the data. Typically, authors combine multiple data sources to formulate a prediction problem. 
We summarize the prediction task information in Table~\ref{tab:mimic-survey}, where: $p_1$ is predicting patient mortality; $p_2$ is patient length of stay; $p_3$ is readmission; $p_4$ is ICD-9 code; $p_5$ is the treatment of a patient; and $p_{+}$ refers to the remaining problems.
Note that $p_{1-3}$ are fully available in \car.

\input{tables/mimic_survey.tex}

Observing Table~\ref{tab:mimic-survey}, we notice that $\sim 30\%$ of the papers tackled the same prediction task ($p_1$, patient mortality).
Given the copious number of prediction problems, we wondered why a generalized framework for reducing repetitive tasks does not exist.
As presented in section~\ref{sec: pred_prob}, we believe we can systematically abstract the problem definition task without losing each model's nuances.
To test our abstraction, we further examined all papers related to $p_1$ to understand how they can be formulated in \car. 

\begin{figure}
    \centering
    \includegraphics[width=1\linewidth]{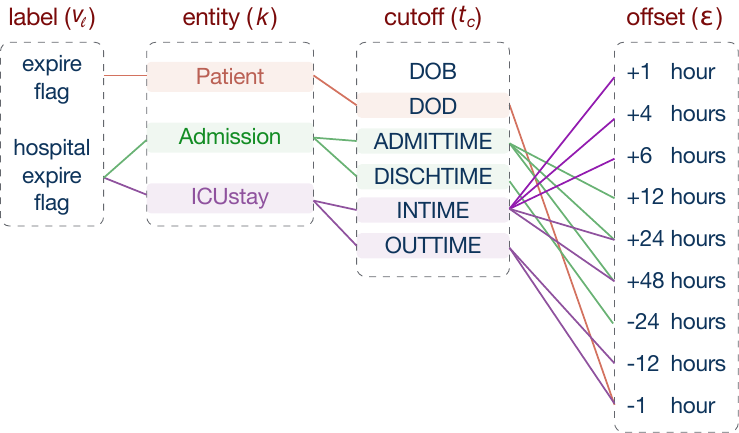}
    \caption{Breakdown of mortality prediction. The enclosing box details how \car's abstraction can articulate all mortality prediction problems presented. The edges showcase all combinations present in $p_1$.}
    \label{fig:mortality_breakdown}
\end{figure}

Figure~\ref{fig:mortality_breakdown} depicts the breakdown of models that solve $p_1$.
Viewing the plot from left to right, we first select \texttt{entityset} components: the target label $v_l$ and the entity of concern $k$.
Next, we identify time-sensitive formulations through $t_c$ and $\epsilon$. 
For example, one of the most common formulations is predicting patient mortality after 48 hours of admission.
In this case, our reference point $t_c$ is set to the time of admission and the offset for learning is $\epsilon=48$ hours.
We discovered that each rendition of $p_1$ surveyed\footnote{Only 22 papers out of 30 mentioned the specifics of their model.} fit our proposed abstraction.

\subsection{Case study: Adaptivity}
\input{tables/classic_cardea.tex}
After defining a widely accepted schema, one possible alternative is to write software for the rest of the process: data assembly, manipulation, pre-processing, and feature engineering. This approach was recently proposed by Wang et al., 2019 for the \mimic database. Their tool, \texttt{MIMIC-Extract}, standardized this process and produced a structured input for widely known prediction tasks - mortality and length of stay~\cite{wang2019mimicextract} - the idea being that other users can exploit this software as long as their data follows the same schema. In addition to requiring subject matter experts with data science expertise to write software, this approach requires significant software maintenance. For instance, \textit{What if a new user only has data corresponding to a certain subset of tables?} \textit{What if the original schema is updated over time?} Software that is hard-coded can be brittle and would need consistent updates. Over time, this process would inevitably lead to large technical debt.

In this case study, we ask whether \car can help mitigate these two situations. Can \car's automated submodules for feature engineering and data assembly do away with extensive hard-coded software engineering? Can it do so while maintaining competitiveness against experts? Even if automation creates competitive solutions, are they expressive? 

\noindent \textbf{Reducing the need for extensive software (re)engineering}: 
\texttt{MIMIC-Extract} extracts 8 tables from the \mimic database to prepare data for processing.
Over 1,000 lines of code were written for the sole purpose of data structuring and featurization.
We ask, how can more data from \mimic be utilized? Can features other than the ones intended be used?

\begin{figure}[ht]
    \centering
    \input{plots/supp_correlation_hist.tex}
    \caption{Histogram of feature correlation using Pearson's correlation coefficient between expert-identified features through \texttt{MIMIC-Extract} and automatic feature generation by \car}
    \label{fig:correlation}
\end{figure}
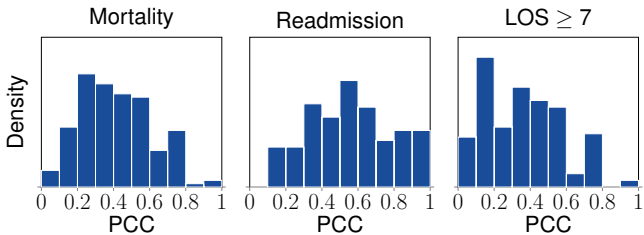

Our proposed framework adapts the \texttt{entityset} to any subschema available, thereby enabling the ingestion of any data type.
Moreover, we use automated feature engineering.
Figure~\ref{fig:correlation} shows the distribution of correlations between expert-generated features and the most correlated feature from automated feature engineering.  \car was able to extract a high proportion of features correlated to \texttt{MIMIC-Extract}. \car also generated features that were not in \texttt{MIMIC-Extract}: \car generated a total of ~400 features, while the latter generated  180 features. 
The median correlation value is 0.39, 0.54, and 0.35 for the mortality, readmission, and length of stay features, respectively.
What are these features, and how significant are they? To investigate this more deeply, we examined the important features (generated using the Random Forest (RF) classifier) which showed that \car relies on variables fetched from different sources, including the number of times the patient was previously diagnosed, the type of admission, and the number of conducted lab tests, which were fetched from the \textit{diagnosis}, \textit{admission}, and \textit{labevents} tables respectively.
On the other hand, the \texttt{MIMIC-Extract} approach focused on utilizing patient readings such as the Glasgow Coma Scale, lactic acid measurements, and arterial pH mean values, which all trace back to the \textit{chartevents} table.

\begin{figure}[!t]
    \vspace{-0.5em}
    \centering
    \input{plots/supp_dimension_lineplot.tex}
    % \vspace{-3pt}
    \caption{The score of the model as we add more data tables. The line represents the smoothed curve of the best-performing classifier.}
    \label{fig:f1_score_max}
    \vspace{-1.2em}
\end{figure}
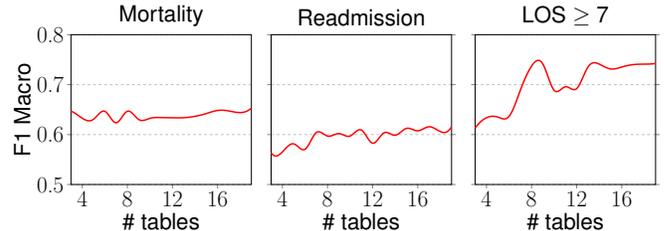

\noindent \textbf{Adapting to data availability}
The automation in \car not only eliminates the need for meticulously engineering features, it also allows for taking advantage of all the data at hand.
But why is more data useful? Are the tables referred to by \texttt{MIMIC-Extract} only the important ones?
To examine this further, we fed \car the same tables as \texttt{MIMIC-Extract}, set it as our baseline, and introduced more tables. From figure~\ref{fig:f1_score_max}, we can see that in certain prediction problems the introduction of more tables improves the model's performance - specifically in the LOS prediction. The best model was able to increase in performance from $0.61$ to $0.74$ (a 21\% improvement).
A huge performance increase happened at $x=6$, as table \texttt{procedureevents} was introduced, which is not included in the \texttt{MIMIC-Extract} pipeline.
But how does the performance of \car compare to \texttt{MIMIC-Extract}?

Table~\ref{tab:classic_vs_cardea_results} summarizes the results obtained from applying \textsc{AutoML} to features that were automatically generated by \car versus \texttt{MIMIC-Extract}, which we consider to exemplify a classical approach to making a prediction model.
While \texttt{MIMIC-Extract}'s approach surpasses \car in mortality prediction by 12\%, the latter triumphs in the other two problems.
Moreover, in comparison to the other classifiers we tested,
(see appendix~\ref{app:manual_features} for all classifiers), 
\car had a more consistent predictive accuracy, indicating that the features it considered are more telling.
Overall, \car showed competitiveness and an ability to extract useful features from any shaped dataset.

%% file: tables/mimic_survey.tex
\begin{table}[!htb]
    \centering
    \begin{threeparttable}
    \begin{tabularx}{\linewidth}{@{}l *{6}{Y} Y}
    \toprule
                                                  & \multicolumn{3}{c}{\car}     & \multicolumn{3}{c}{External} & \\ 
                                                     \cmidrule(lr){2-4}                   \cmidrule(lr){5-7}
                                                  & $p_1$        & $p_2$           & $p_3$       
                                                  & $p_4$        & $p_5$           & $p_{+}$           & Total\\
                                    % &              & Mortality    & Length of Stay  & Readmission       
                                    %               & ICD-9 Code   & Treatment       & Other              \\
    % \multirow{3}[3]{*}{\begin{sideways}Data Structure\end{sideways}}
    \midrule
                                     Tabular      &     19       &     6           &    3            & 
                                                        7        &     6           &    8            & 49 \\ \midrule
                                     Time series  &     6        &     1           &    0            &
                                                        1        &     1           &    13           & 22 \\ \midrule
                                     Unstructured &     5        &     0           &    1            &
                                                        15       &     3           &    7            & 31 \\ 
    \midrule
                                    Total         &     30       &     7           &    4            &
                                                        23       &     10          &    28           & 102\tnote{*} \\
    \bottomrule
    \end{tabularx}
    \begin{tablenotes}
      \item[*] Papers targeting multiple prediction problems / using multiple data structures are counted separately.
    \end{tablenotes}
    \end{threeparttable}
    \caption{\mimic Surveyed Papers Summary.}
    \label{tab:mimic-survey}
    % \vspace{-0.5cm}
\end{table}

%% file: tables/classic_cardea.tex
\begin{table*}[!htb]
    \centering
    \begin{tabularx}{\linewidth}{@{} *{1}{l} *{6}{Y}}
    \toprule
         &  \multicolumn{3}{c}{\texttt{MIMIC-Extract} + \textsc{AutoML}} & \multicolumn{3}{c}{\car} \\
        %  &  \multicolumn{2}{c}{Classic vs. \car}\\
                \cmidrule(lr){2-4} \cmidrule(lr){5-7} 
        % \cmidrule(lr){8-9} 
         & $\mu$ $\pm$ $\sigma$     &  \textit{best} & \textit{best CL}                       &  $\mu$ $\pm$ $\sigma$  & \textit{best}  & \textit{best CL} \\
        %  &  \textit{p-value} & \textit{p-value} \\
    \midrule
    Mortality   & 0.540 $\pm$ 0.1643 & 0.740  & LR   & 0.566 $\pm$ 0.0529 & 0.660  & XGB    \\
    Readmission & 0.459 $\pm$ 0.1455 & 0.540  & SGD  & 0.540 $\pm$ 0.0628 & 0.635  & XGB    \\
    LOS $\geq$ 7& 0.560 $\pm$ 0.1296 & 0.691  & GB   & 0.519 $\pm$ 0.1650 & 0.789  & LR     \\
    \bottomrule
    \end{tabularx}
    \vspace{1pt}
    \caption{Comparison of F1 Macro scores for models whose features were generated by classical approaches (\texttt{MIMIC-Extract}) and models whose features were generated automatically (\car)}
    \label{tab:classic_vs_cardea_results}
\end{table*}

%% file: plots/supp_correlation_hist.tex
    \begin{tikzpicture}[scale=0.35,font=\sffamily\sansmath]
    \begin{axis}[
    title={Mortality},title style={font=\sffamily\YUGE},
    ylabel={Density},
    xlabel={PCC},
    ybar,
    ymin=0,ymax=45,
    xmin=0,xmax=1,
    x label style={font=\sffamily\YUGE},y label style={font=\sffamily\YUGE},
    tick align = outside,xtick pos=left, ticklabel style={font=\Huge}, ytick=0, yticklabels={,,}
    ]
    \addplot+[hist={bins=10,data min=0.0,data max=1},fill={rgb:red,1;green,3;blue,6},draw=white] table [y index = 0] {data/hist_mortality.csv};
    \end{axis}
    \end{tikzpicture}
    \begin{tikzpicture}[scale=0.35,font=\sffamily\sansmath]
    \begin{axis}[
    title={Readmission},title style={font=\sffamily\YUGE},
    xlabel={PCC},
    ybar,
    ymin=0,ymax=45,
    xmin=0,xmax=1,
    x label style={font=\sffamily\YUGE},y label style={font=\sffamily\YUGE},
    tick align = outside,xtick pos=left, ticklabel style={font=\Huge}, ytick=0, yticklabels={,,}
    ]
    \addplot+[hist={bins=10,data min=0.0,data max=1},fill={rgb:red,1;green,3;blue,6},draw=white] table [y index = 0] {data/hist_readmission.csv};
    \end{axis}
    \end{tikzpicture}
    \begin{tikzpicture}[scale=0.35,font=\sffamily\sansmath]
    \begin{axis}[
    title={LOS $\geq$ 7},title style={font=\sffamily\YUGE},
    xlabel={PCC},
    ybar,
    ymin=0,ymax=45,
    xmin=0,xmax=1,
    x label style={font=\sffamily\YUGE},y label style={font=\sffamily\YUGE},
    tick align = outside,xtick pos=left, ticklabel style={font=\Huge}, ytick=0, yticklabels={,,}
    ]
    \addplot+[hist={bins=10,data min=0.0,data max=1},fill={rgb:red,1;green,3;blue,6},draw=white] table [y index = 0] {data/hist_los.csv};
    \end{axis}
    \end{tikzpicture}

%% file: plots/supp_dimension_lineplot.tex
\begin{tikzpicture}[scale=0.35,font=\sffamily\sansmath] 
\begin{axis}[
title={Mortality},title style={font=\sffamily\YUGE},
xlabel={\# tables},
ylabel={F1 Macro},
xmin=3,xmax=19,
ymin=0.5,ymax=0.8,
x label style={font=\sffamily\YUGE},y label style={font=\sffamily\YUGE},
y tick label style={/pgf/number format/.cd,fixed,fixed zerofill,precision=1,1000 sep={}},
ymajorgrids=true,grid style=dashed,
tick align = outside,xtick pos=left,ytick pos=left, ticklabel style={font=\Huge}, xtick={4,8,12,16}
]
\addplot[color=red,mark=none,solid,ultra thick] table[x = x,y = y, col sep=comma] {data/dim_mortality.csv};
\end{axis}
\end{tikzpicture}
\begin{tikzpicture}[scale=0.35,font=\sffamily\sansmath] 
\begin{axis}[
title={Readmission},title style={font=\sffamily\YUGE},
xlabel={\# tables},
xmin=3,xmax=19,
ymin=0.5,ymax=0.8,
x label style={font=\sffamily\YUGE},y label style={font=\sffamily\YUGE},
y tick label style={/pgf/number format/.cd,fixed,fixed zerofill,precision=1,1000 sep={}},
ymajorgrids=true,grid style=dashed,
tick align = outside,xtick pos=left, xticklabel style={font=\Huge}, yticklabels={,,}, xtick={4,8,12,16}
]
\addplot[color=red,mark=none,solid,ultra thick] table[x = x,y = y, col sep=comma] {data/dim_readmission.csv};
\end{axis}
\end{tikzpicture}
\begin{tikzpicture}[scale=0.35,font=\sffamily\sansmath] 
\begin{axis}[
title={LOS $\geq$ 7},title style={font=\sffamily\YUGE},
xlabel={\# tables},
xmin=3,xmax=19,
ymin=0.5,ymax=0.8,
x label style={font=\sffamily\YUGE},y label style={font=\sffamily\YUGE},
y tick label style={/pgf/number format/.cd,fixed,fixed zerofill,precision=1,1000 sep={}},
ymajorgrids=true,grid style=dashed,
tick align = outside,xtick pos=left, xticklabel style={font=\Huge}, yticklabels={,,}, xtick={4,8,12,16}
]
\addplot[color=red,mark=none,solid,ultra thick] table[x = x,y = y, col sep=comma] {data/dim_los.csv};
\end{axis}
\end{tikzpicture}

%% file: manuscript/userstudy.tex
\section{User Study}
\label{sec:userstudy}
We conducted a user study to evaluate \car~from different angles and to answer the following questions:
(\textbf{R1}) \textit{Can users understand the concepts and functions of \car~quickly and correctly? } 
(\textbf{R2}) \textit{How effective is \car in supporting medical data analysis?
and} (\textbf{R3}) \textit{How do users perceive the usefulness of \car?} 

% \subsection{Participants and apparatus}
We held a 90-minute workshop with 10 participants, 7 males and 3 females, each having between 4 and 12 years of coding experience ($\mu$=7.7, $\sigma$=2.53) and between 2 and 5 years of machine learning experience ($\mu$=4.0, $\sigma$=1.10).
The participants included graduate students, data analysts, researchers, and engineers. They are general users whose daily work involves data analysis, but they are not experts in analyzing medical data.

\subsection{Tasks and Procedure}

We started the workshop with a 30-minute tutorial+demo session, introducing the relevant background knowledge and demonstrating how to use \car. Next, we set up an one-hour quiz. We created our study datasets based on the \textsc{Mimic-iii} Demo data~\footnote{\url{https://physionet.org/content/mimiciii-demo/}}. The sampled data contains the same tables as the complete dataset, but the number of patients is reduced to 100.

The quiz required the participants to use \car to perform four tasks.
Task 1 asked the participants to use \car~ to load a specified dataset and observe the loaded tables. Task 2 asked the participants to explore potential prediction problems and then use \car~to explore the predefined problems. Task 3 required the participants to use \car~ to generate the feature matrix and prepare training and testing data for later use. Task 4 requested the participants to explore different pipelines and report the accuracy scores of the obtained models. 
To investigate whether users can effectively understand and use \car~ (\textbf{R1}), after every task, the participants were asked several questions (Table~\ref{tab:tasks_merge}). 
We recorded the task accuracy and completion time. We also asked the participants to estimate what the completion time sould be without \car~(\textbf{R2}). Four options ($<10$ minutes, $<1$ hour, $<3$ hours, and $\geq3$ hours) were given, and we used $5$, $35$, $120$, and $180$ respectively as the actual estimated times in order to compute the average time. 
%The comparison between the estimated time and actual task completion time can provide insights into the effectiveness of \car (\textbf{R2}).  

At the end of the quiz, to get a more comprehensive understanding of \car~'s usefulness (\textbf{R3}), we asked the participants to fill out a survey to rate \car. We used 5-point Likert-scale questions with 1 being "strongly disagree" and 5 being "strongly agree."
The questions covered ease of use, ease of learning, effectiveness in solving medical data prediction problems, satisfaction, and utility (i.e., whether it provides all the features you need.).

\input{tables/tasks_merge.tex}

\input{tables/overall_ratings.tex}

\subsection{Result Analysis}
% \subsection{Quiz Result Analysis}
\noindent{\textbf{Quiz results---task accuracy}}
% \textbf{Task accuracy.}
% \noindent \textbf{Reducing the need for extensive software (re)engineering}: 
Table~\ref{tab:tasks_merge} summarizes the quiz results. We observed that all of the participants were able to answer the three questions (Q1, Q7, and Q8) related to data assembler (T1), featurizer (T3), and ML pipelines (T4), respectively.
However, a few participants still failed to answer Q3 (with T2---problem definer; acc$=70\%$, 7/10) and Q6 (with T3---featurizer; acc$=80\%$, 8/10). 
This indicates that the participants had a good understanding of how data were loaded, transformed (y labels) and fed for training predictive models.
Three participants (1 No and 2 Not sure) had difficulty  judging whether a prediction problem was feasible given the data. This further motivated us to automatically provide a comprehensive prediction problem list.
We also found that two participants had difficulty in understanding the feature matrix (Q6). 

We asked several other questions to collect feedback from the participants, including Q2, Q4, Q5, and Q9. 
For Q2, the participants rated helpfulness on a 5-point Likert-scale score ($\mu$=4.0, $\sigma$=1.00). The prevailing score, 4, suggested that most people (70\%, 7/10) thought this feature very helpful.
For both Q4 and Q5, our purpose was to investigate whether our predefined problem list could satisfy user demands. 
The results reported that 4 out of 10 (40\%) participants found their target prediction problems existed in Cardea, while the rest thought of a question outside.
For Q9, the average accuracy is 89.8\%, with the standard deviation being 4.30\%, which suggests that the participants were able to try different pipelines to run the experiments. We further noted that one participant obtained 100\% accuracy, which was possible as the experiments were run on a sampled dataset.

\noindent{\textbf{Quiz results---task completion time.}}
From table~\ref{tab:tasks_merge}, we found that using \car~ can significantly improve the efficiency of performing each type of task. The speedup time ranged from 7 to 27, where T3 (featurizer) had the largest efficiency increase (27x). On average, tasks regarding the types were expected to be completed in around 4 to 6 minutes using \car. T2 ($\mu$=5.77 min, $\sigma$=2.11) took the longest; this could be because participants had to spend time observing the metadata and investigating the problems. Although T4 ($\mu$=4.25 min, $\sigma$=3.77) had the least completion time, its SD was the highest. This can be explained by the fact that users can generally use \car~ to learn a model in a short time, and the time a user spent depends on his/her expectation regarding performance.  

% \subsection{Survey Result Analysis}
\noindent{\textbf{Survey results---usability.}}
Overall, as shown in Table~\ref{tab:overall_ratings}, we received good feedback from the participants.
% Overall, we received good feedback from the participants.
Most participants agree or strongly agree that \car~ is easy to use (9/10, $\mu$=4.2, $\sigma$=0.6), easy to learn (8/10, $\mu$=4.1, $\sigma$=0.7), and able to significantly improve their efficiency to perform prediction tasks (7/10, $\mu$=4.2, $\sigma$=0.87). Few participants held neutral options on these three aspects, which confirmed with our quiz results that few participants had difficulty in identifying proper prediction problems and understanding the auto-featurization process. Nearly all participants (8/10, $\mu$=4.2, $\sigma$=0.75) felt satisfied with the presented tool. 
% and willing to use this tool to perform predictive tasks for medical data in the future. 
Compared with the other four, the ``utility'' was scored relatively lower ($\mu$=3.8, $\sigma$=0.87), suggesting the absence of some user-desired features.  
The concerns lay mainly in two aspects: visual interface and controllability. One participant pointed out that a friendly visual interface would make \car~ accessible to a wider audience. 
% One other comment suggested that visualizing some key information, such as table relationships or system inputs/outputs, would be helpful to understand the workflow. 
% the framework could improve its usability by integrating a visual aesthetic, which would make it accessible to a wider audience.
% One participant suggested visualizing the relationships of the tables, especially when such relationships are highly complicated. 
% Several participants mentioned that such a few number of predefined prediction problems could not meet his/her demands. 
As for the controllability, 
one critic was concerned about high automation partly limiting users' ability to explore freely. 
% \hl{For example, the current schema of ``Feature'' objects returned by Featuretools only allows for a high-level observation on the feature matrix and cannot watch feature-level information or perform feature-level operations}
% \sarah{I think what was meant here is that, the featurization component only returns the feature matrix, featuretools also returns the name of the generated features, but it was not shown in the example} \hl{(e.g., setting weighted samples)} \sarah{Also I think this is not part of featurization, but modeling. we should have a preprocessing primitive that would do that}.
% One also commented that prompting whether a ML model is suitable for a specified prediction problem would be helpful. \hl{See more from the survey responses.}

% We also collected additional feedback from the participants in the survey. One of the main areas of improvement suggested was that, the framework could improve its usability if it were accompanied by a visual aesthetic, which would make it accessible to a wider audience. \hl{Give some real comments from users here. Xx commended, ``balala''. Xx praised, ``balala''. Xx mentioned, ``balala''. Xx suggested, ``balala''.}

% \input{tables/tasks.tex}
% \input{tables/tasks_data.tex}

%% file: tables/tasks_merge.tex
\begin{table*}
  \centering
  \setlength{\extrarowheight}{0pt}
  \addtolength{\extrarowheight}{\aboverulesep}
  \addtolength{\extrarowheight}{\belowrulesep}
  \setlength{\aboverulesep}{0pt}
  \setlength{\belowrulesep}{0pt}
  \begin{tabular}{c|l|c|ccc} 
    \toprule
    \multicolumn{1}{l|}{ \textbf{Task} } & \multicolumn{1}{c|}{\textbf{Question} }                                                                                                                                                                                                                                                                                                                                                                                                                                                                                                                                                                                                                                                                                                                                                                                                                                                                                                                                                                & \multicolumn{1}{l|}{\textbf{Accuracy} }               & \multicolumn{3}{c}{\textbf{Completion Time} }                    \\ \cmidrule(lr){4-6} 
                                         &                                                                                                                                                                                                                                                                                                                                                                                                                                                                                                                                                                                                                                                                                                                                                                                                                                                                                                                                                                                              &                                                       & Cardea (min)          & Estimated (min)          & Speedup Times ($\simeq$x)        \\ 
    % \cmidrule{1-3}\cmidrule(lr){4-4}\cmidrule(lr){5-5}\cmidrule(lr){6-6}
    \midrule
    T1                                   & \begin{tabular}[c]{@{}l@{}}Q1.~\textcolor[rgb]{0.122,0.122,0.141}{How many entries were in the "icustays" table?}\\\textcolor[rgb]{0.122,0.122,0.141}{\textcolor[rgb]{0.122,0.122,0.141}{\textcolor[rgb]{0.122,0.122,0.141}{Q2.~}Is it helpful to display the "relationship" of the entityset?\textcolor[rgb]{0.122,0.122,0.141}{}} }\end{tabular}                                                                                                                                                                                                                                                                                                                                                                                                                                                                                                                                                                                                                                                     & \begin{tabular}[c]{@{}c@{}}100\%\\~ \\ \end{tabular} & 4.86 (2.25)           & 57.50 (57.54)           & 12            \\ 
    \midrule
    T2                                   & \begin{tabular}[c]{@{}l@{}}Q3.~\textcolor[rgb]{0.122,0.122,0.141}{Given the data, can we predict a patient's ICU time?}\\\textcolor[rgb]{0.122,0.122,0.141}{\textcolor[rgb]{0.122,0.122,0.141}{\textcolor[rgb]{0.122,0.122,0.141}{\textcolor[rgb]{0.122,0.122,0.141}{Q4.~}\textcolor[rgb]{0.122,0.122,0.141}{Given the data, what is your prediction problem?}}}}\\\textcolor[rgb]{0.122,0.122,0.141}{\textcolor[rgb]{0.122,0.122,0.141}{\textcolor[rgb]{0.122,0.122,0.141}{\textcolor[rgb]{0.122,0.122,0.141}{\textcolor[rgb]{0.122,0.122,0.141}{\textcolor[rgb]{0.122,0.122,0.141}{\textcolor[rgb]{0.122,0.122,0.141}{\textcolor[rgb]{0.122,0.122,0.141}{\textcolor[rgb]{0.122,0.122,0.141}{\textcolor[rgb]{0.122,0.122,0.141}{\textcolor[rgb]{0.122,0.122,0.141}{\textcolor[rgb]{0.122,0.122,0.141}{Q5.~}\textcolor[rgb]{0.122,0.122,0.141}{Does your prediction problem already exist in Cardea?}\textcolor[rgb]{0.122,0.122,0.141}{}\textcolor[rgb]{0.122,0.122,0.141}{}}}}}}} }}}}}\end{tabular} & \begin{tabular}[c]{@{}c@{}}70\%\\ ~ \\ ~ \\ \end{tabular} & \textbf{5.77 (2.11)}  & 40.00 (42.19)           & 7          \\ 
    \midrule
    T3                                   & \begin{tabular}[c]{@{}l@{}}Q6.~\textcolor[rgb]{0.122,0.122,0.141}{How many features were generated?}\\\textcolor[rgb]{0.122,0.122,0.141}{\textcolor[rgb]{0.122,0.122,0.141}{\textcolor[rgb]{0.122,0.122,0.141}{Q7.~}\textcolor[rgb]{0.122,0.122,0.141}{What is the datatype of y?}\textcolor[rgb]{0.122,0.122,0.141}{}} }\end{tabular}                                                                                                                                                                                                                                                                                                                                                                                                                                                                                                                                                                                                                                                                 & \begin{tabular}[c]{@{}c@{}}80\%\\100\% \end{tabular}  & 4.45 (0.84)           & \textbf{121.5 (67.57)}  & \textbf{27}   \\ 
    \midrule
    T4                                   & \begin{tabular}[c]{@{}l@{}}\textcolor[rgb]{0.122,0.122,0.141}{Q8. Can you get a model with accuracy \textgreater{} 80\%?}\\\textcolor[rgb]{0.122,0.122,0.141}{\textcolor[rgb]{0.122,0.122,0.141}{}Q9. What was the best accuracy score you obtained?\textcolor[rgb]{0.122,0.122,0.141}{} }\end{tabular}                                                                                                                                                                                                                                                                                                                                                                                                                                                                                                                                                                                                                                                                                                & \begin{tabular}[c]{@{}c@{}}100\% \\ ~ \\ \end{tabular}  & 4.25 (3.77)           & 55.63 (51.20)           & 13            \\
    \bottomrule
  \end{tabular}
  \vspace{4pt}
  \caption{Experimental tasks, questions, and results.
  Tasks are categorized into four classes corresponding to Cardea's 4 components: data assembler (T1), problem definer (T2), featurizer (T3), and ML pipelines (T4).  Task completion time is shown in Avg. (SD). The speedup times of every task are reported.}
  \label{tab:tasks_merge}
  \vspace{-2em}
\end{table*}

%% file: tables/overall_ratings.tex
\begin{table}
  \centering
  \setlength{\extrarowheight}{0pt}
  \addtolength{\extrarowheight}{\aboverulesep}
  \addtolength{\extrarowheight}{\belowrulesep}
  \setlength{\aboverulesep}{0pt}
  \setlength{\belowrulesep}{0pt}

  \begin{tabular}{p{6.5cm}cc}
  \toprule
  Ratings on                                         & Mean & SD    \\ 
  \cmidrule(lr){1-1}\cmidrule(lr){2-2}\cmidrule(lr){3-3}
  Ease of use                                        & 4.2  & 0.6   \\
  \rowcolor[rgb]{0.933,0.933,0.933} Ease of learning & 4.1  & 0.7   \\
  Effectiveness (in medical data predictive analysis)& 4.2  & 0.87  \\
  \rowcolor[rgb]{0.933,0.933,0.933} Satisfaction     & 4.2  & 0.75  \\
  Utility (cover all desired features)               & 3.8  & 0.87  \\
  \bottomrule
  \end{tabular}
  \vspace{1pt}
  \caption{Ratings on the overall experience on \car~(1 $=$ strongly disagree, 2 $=$ disagree,
  3 $=$ neutral, 4 $=$ agree, 5 $=$ strongly agree).}
  \label{tab:overall_ratings}
  \vspace{-20pt}
\end{table}

%% file: manuscript/discussion.tex
\section{Further Discussion}
\label{sec:discussion}
\car takes \textsc{AutoML} to the next level by engineering components that allow for flexible data ingestion, problem selection, and data- and model-understanding. In this section, we shed light on some of the more interesting findings from the results obtained in section~\ref{sec:results}.

\noindent \textbf{Adaptivity of Automatic Data Assembler}:
Most publicly available EHR datasets involve minimal, non-sensitive information about patients, and do not represent real scenarios. Therefore, we needed to test the robustness of the automatic data assembler. For this, we randomly sampled 100 subsets of \mimic and fed it to the framework. In all cases, the framework adapts to the dimensions of the data and seamlessly proceeds to the next components.

\noindent \textbf{Acceleration in Building Prediction Models}:
We conduct another experiment where we try to solve a new problem; predicting the risk of acquiring a certain disease~\cite{cheng2016risk}. We utilize a large dataset of $1.1$ million encounter records from a private EHR entity. The dataset is composed of 10 separate hospital records that were merged together. Using \car, not only were they able to run the framework from raw data to deployable models, but they achieved this result in $< 1$ month.

\noindent \textbf{Model Interpretation}: The data- and model-auditor supplement the user with vital information describing the model.
Continuing on the same real-world application mentioned, we leveraged the property of auditing into understanding the characteristics of the data. For example, out of the collection of 10 hospitals, we noticed that $hospital_1$ performed better than $hospital_2$, because the patient cohort in $hospital_1$ is less diverse than in $hospital_2$, decreasing the variance of the training data~\cite{suresh2018learning}.

%% file: manuscript/conclusion.tex
\section{Concluding Remarks}
\label{sec:conclusion}

In this paper, we introduced \car, an open framework for creating healthcare prediction models automatically. In \car, we adopt HL7's FHIR standard as an interface for representing data, and provide an abstraction for defining prediction problems, a set of coded prediction problems, and multiple modeling approaches. Through the proposed design, users overcome the limitations of manual feature engineering and single-model application, and are not limited to one prediction problem. While automation allows for scaling the number of prediction problems and hospitals that we can address without much manual effort, it also allows us to focus on developing a comprehensive systematic approach for assessing data and models. 
We demonstrated the efficacy of this framework by solving 5 prediction problems, comparing its performance to existing models, and evaluating the usability by a user study.

% \section*{Acknowledgement}
% This work was supported by a grant from the Center for Complex Engineering Systems (CCES) at King Abdulaziz City for Science and Technology (KACST) and Data to AI lab at Massachusetts Institute of Technology (MIT). 

%% file: manuscript/supplementary_material.tex
\subsection{Auditing}
\label{auditing}

\subsubsection{Data Auditor}
Data from different hospitals will inevitably have variations even if they conform to a standardized schema\footnote{Note that data curation and cleaning -- in other words, standardizing field names or formats -- does not apply here, because the data already conforms to the FHIR standard.}. 
These variations stem from hospitals' differing execution of health management systems, resulting, for example, in missing values for certain fields; serving different populations.
We most commonly encountered lack of demographic variables and diagnosis information; data from different and/or short time periods (temporal coverage); data from different geographical areas (spatial coverage); and general quality issues (inaccuracy, incompleteness, format inconsistency, etc.).  

In a typical data science endeavor, data scientists summarize these caveats; while they don't preclude the building of a predictive model, they do affect the model's accuracy. 
For example, imagine a prediction problem for which \texttt{age} is an important feature. 
A dataset with incomplete age-related data will reduce the overall accuracy of the model. 
These issues should be reported in a systematic fashion, enabling the user to debug and ultimately trust the models. 
To address this, we created a \textit{data auditor}. The data auditor uses a dictionary of checks for different fields and generates a data summary report. 
These checks are divided into two categories: data quality metrics and distribution checks. 

\noindent \textbf{Data quality metrics}: calculate metrics pertaining to missing information in the entityset.
One example is the number and percentage of missing values (i.e. \texttt{NaN} values). \\
\noindent \textbf{Data distribution checks}: consider if the entityset complies with \textit{general} assumptions regarding the data. 
These checks record the distribution, number of unique values for categorical values, percentage frequency, and the minimum/maximum values, all of which are compared against standard expected values. 
% \footnote{The library is publicly open for all users on github: \url{https://github.com/DAI-Lab/Cardea}.}

\subsubsection{Model Auditor} 
The ability to audit and report a prediction model's result is critical, as it allows domain experts to assess a trained model's performance. Due to variations in problem type (regression, binary, or multi-classification), issues with data imbalance, sparse or high-dimensional feature matrices, and different applications of models, it is difficult to find a single metric applicable across different prediction problems and algorithms. However, a set of metrics can be identified that provide an overall understanding for various prediction problems while simplifying models' complexities. Furthermore, the model auditor allows users to apply their own metrics to the model's results. We provide evaluation metrics like accuracy, F1 scores, precision, recall, and area under the curve (AUC) (for classification problems); and mean square errors, mean absolute errors, and R squared (for regression problems). 

\subsection{Datasets}
\label{app:datasets}
\kaggle, a platform where data scientists compete to solve various prediction problems ~\cite{kaggleoverview}, presented a competition in May 2016 to predict patient no-show. 
The dataset is is composed of 110,527 labeled scheduled appointments. Each appointment is described by a limited patient demographic information, the name of the physician the patient is booked with, the dates in which the appointment was scheduled and the actual date of the appointment, and a flag operator that mentions whether the patient has received a reminder or not.

\begin{table}[ht]
    \centering
    \begin{tabularx}{\linewidth}{lc}
    \toprule
  
                                    & Appointment Data\\
                                    \midrule

    \textsc{Data assembler}               &\\    
    \hspace{2em} No. of loaded variables         & 21\\
    \hspace{2em} No. of loaded resources         & 9\\
    \textsc{Appointment no-show}    &\\
    \hspace{2em} No. training examples           & 110,527 \\
    \hspace{2em} Positive classes ratio (\%)     & 22,319 (20.19\%)	\\
    \hspace{2em} Negative classes ratio (\%)     & 88,208	(79.81\%)  \\
    \textsc{Featurization}          &\\   
    \hspace{2em} No. generated features          & 95 \\
    \bottomrule
    \end{tabularx}
    \vspace{1pt}
    \caption{Metadata report created by \car.}
    \label{kaggle_data}
\end{table}

\begin{table}[ht]
  \begin{tabularx}{\linewidth}{XXX}
    \toprule
                                    & Negative        & Positive\\
    \midrule
    Scholarship                      & 90.71\%        & 9.29\%\\
    Hypertension                     & 80.35\%        & 19.65\%\\
    Diabetes                         & 92.91\%        & 7.09\%\\
    Alcoholism                       & 97.58\%        & 2.42\%\\
  \bottomrule
  \end{tabularx}
  \vspace{1pt}
  \caption{Cohort of \kaggle generated by the data auditor.}
  \label{kaggle_data_audit}
\end{table}

\input{tables/mimic_results_regression.tex}
\input{tables/mimic_metadata.tex}
\input{tables/mimic_extract_results.tex}
\input{tables/mimic_results_classification.tex}

\mimic is a de-identified, freely accessible relational EHR database from Beth Israel Deaconess Medical Center in Boston, Massachusetts. The dataset spans patients who stayed within the Intensive Care Unit (ICU) between 2001 and 2012.
It provides multiple attributes: demographics, vital signs, laboratory tests, medical notes, etc. Overall, there are approximately $\sim 59K$ hospital admissions for $38,597$ distinct patients detailed in \mimic.

Table~\ref{mimic_data} shows the number of training examples, the label split, and number of features generated across the different problems defined in the use-cases.

\subsection{Features and Model Training}
\label{app:manual_features}
\car produced varied number of features depending on the data ingested, as described in section~\ref{sec: Auto Machine Learning}. The \texttt{MIMIC-Extract} approach ~\cite{wang2019mimicextract} created 180 features; where 10 were from patient and admission details: gender, ethnicity, age, insurance, admittime, dischtime, intime, outtime, admission type, and first careunit. The remaining 170 features were generated from in-hospital information:  \textit{chartevents} and \textit{labevents}.

\noindent \textbf{Model Training}: models were training using 80-20 train and test split. We used a 10 fold cross validation scheme. We use Bayesian hyperparamter tuning and performed 100 evaluations. In addition, we performed bootstrapping to measure the robustness of the results, which is shown by the confidence interval posed in table~\ref{mimic_classification_results}.

%% file: tables/mimic_results_regression.tex
\begin{table}
\centering
\begin{tabular}{llcc}
\toprule
              & {} & \multicolumn{2}{c}{LOS} \\
    \cmidrule(lr){3-4}
    Algorithm & {} &  $R^2$ & MSE \\
\midrule
\multirow{2}{*}{Stochastic Gradient Descent} & NT &   0.283 &       0.402 \\
    & T &  0.273 &       0.407 \\
\cline{1-4}
\multirow{2}{*}{XGB} & NT &  0.514 &       0.279 \\
    & T &   \textbf{0.515} &       \textbf{0.278} \\
\cline{1-4}
\multirow{2}{*}{K-Nearest Neighbors} & NT &  0.269 &       0.409 \\
    & T &  0.258 &       0.416 \\
\cline{1-4}
\multirow{2}{*}{Ridge} & NT &   0.372 &       0.355 \\
    & T &   0.373 &       0.354 \\
\cline{1-4}
\multirow{2}{*}{AdaBoost} & NT &  -0.156 &       0.64 \\
    & T  & -9.007 &       5.337 \\
\cline{1-4}
\multirow{2}{*}{SVR} & NT &   -1.113 &       1.142 \\
    & T &  -0.733 &       0.944 \\
\cline{1-4}
\multirow{2}{*}{Linear Regression} & NT &    -1.1e+20 &       4.9e+19 \\
    & T &  -1.6e+17 &       7.5e+16 \\
\bottomrule
\end{tabular}
\vspace{4pt}
% \captionof{table}{\car regression performance on \mimic.} 
\caption{Results from predicting the number of days a patient will stay at the hospital (regression) using \mimic.}
\label{tab:mimic_regression_results}
\end{table}

%% file: tables/mimic_metadata.tex
\begin{table*}[ht]
    \centering
    \begin{tabular}{lccc}
    \toprule
  
                                    & Mortality & Readmission & LOS $\geq$ 7 \\
                                    \midrule
    \textsc{Problem Definition}    &\\
    \hspace{2em} No. training examples           &  58,928             & 58,464            & 61,478            \\
    \hspace{2em} Positive classes ratio (\%)     &  5,849 (9.92\%)     & 3,406 (5.83\%)	& 9,872 (16.06\%)   \\
    \hspace{2em} Negative classes ratio (\%)     &  53,079 (90.08\%)   & 55,058 (94.17\%)  & 51,606 (83.94\%)  \\
    \textsc{Featurization}          &\\   
    \hspace{2em} No. generated features          &  449                & 443               &   178             \\
  \bottomrule
    \end{tabular}
    \vspace{4pt}
    \caption{Metadata report created by \car of \mimic.}
    \label{mimic_data}
\end{table*}

%% file: tables/mimic_extract_results.tex
\begin{table*}[!ht]
    \centering
    \begin{tabular}{lcccccc}
    \toprule
         &  \multicolumn{2}{c}{Mortality} & \multicolumn{2}{c}{Readmission} & \multicolumn{2}{c}{LOS $\geq$ 7} \\
            \cmidrule(lr){2-3} \cmidrule(lr){4-5} \cmidrule(lr){5-7}
         &  F1 Macro & Accuracy & F1 Macro & Accuracy &  F1 Macro & Accuracy \\
    \midrule
    Logistic Regression (LR)         &   0.740 $\pm$ 0.0084   &   0.925 $\pm$ 0.0036 &   0.532 $\pm$ 0.0091   &   0.943 $\pm$ 0.0026  &   0.637 $\pm$ 0.0074   &   0.642 $\pm$ 0.0074   \\
    K-Nearest Neighbors (KNN)        &   0.557 $\pm$ 0.0062   &   0.897 $\pm$ 0.0037 &   0.530 $\pm$ 0.0082   &   0.940 $\pm$ 0.0021  &   0.596 $\pm$ 0.0075   &   0.597 $\pm$ 0.0074   \\
    Random Forest (RF)               &   0.490 $\pm$ 0.0040   &   0.895 $\pm$ 0.0037 &   0.485 $\pm$ 0.0027   &   0.941 $\pm$ 0.0027  &   0.651 $\pm$ 0.0084   &   0.654 $\pm$ 0.0084  \\
    Gaussian Naive Bayes (GNB)       &   0.227 $\pm$ 0.0406   &   0.228 $\pm$ 0.0406 &   0.103 $\pm$ 0.0050   &   0.103 $\pm$ 0.0049  &   0.406 $\pm$ 0.0210   &   0.505 $\pm$ 0.0114   \\
    Multinomial Naive Bayes (MNB)    &   0.474 $\pm$ 0.0015   &   0.893 $\pm$ 0.0036 &   0.485 $\pm$ 0.0027   &   0.941 $\pm$ 0.0027  &   0.549 $\pm$ 0.0064   &   0.586 $\pm$ 0.0048   \\
    XGBoost (XGB)                    &   0.472 $\pm$ 0.0010   &   0.893 $\pm$ 0.0036 &   0.485 $\pm$ 0.0007   &   0.941 $\pm$ 0.0027  &   0.322 $\pm$ 0.0018   &   0.475 $\pm$ 0.0040   \\
    Stochastic Gradient Descent (SGD)&   0.699 $\pm$ 0.0373   &   0.920 $\pm$ 0.0060 &   0.540 $\pm$ 0.0085   &   0.943 $\pm$ 0.0026  &   0.627 $\pm$ 0.0169   &   0.636 $\pm$ 0.0120   \\
    Gradient Boosting (GB)           &   0.658 $\pm$ 0.0118   &   0.915 $\pm$ 0.0025 &   0.509 $\pm$ 0.0078   &   0.942 $\pm$ 0.0027  &   0.691 $\pm$ 0.0094   &   0.693 $\pm$ 0.0093   \\
    \bottomrule
    \end{tabular}
    \vspace{1pt}
    \caption{Results from applying \car's \textsc{AutoML} on several classification problems using \texttt{MIMIC-Extract} features.}
    \label{tab:mimic_extract_results}
\end{table*}

%% file: tables/mimic_results_classification.tex
\begin{table*}[htb!]
\begin{tabular}{llccccccc}
\toprule
              &            & \multicolumn{2}{c}{Mortality} & \multicolumn{2}{c}{Readmission} & \multicolumn{2}{c}{LOS $\geq$ 7} \\
                             \cmidrule(lr){3-4} \cmidrule(lr){5-6} \cmidrule(lr){7-8}
              &            &           F1 Macro &         Accuracy &         F1 Macro &         Accuracy &         F1 Macro &         Accuracy \\
\midrule
\multirow{2}{*}{Logistic Regression} & NT &    0.559 $\pm$ 0.0069 &  0.903 $\pm$ 0.0024 &  0.601 $\pm$ 0.0091 &  0.946 $\pm$ 0.0018 &  0.789 $\pm$ 0.1155 &  0.921 $\pm$ 0.0022 \\
              & T &    0.605 $\pm$ 0.0042 &  0.731 $\pm$ 0.0041 &  0.601 $\pm$ 0.0090 &  0.946 $\pm$ 0.0017 &  0.789 $\pm$ 0.1155 &  0.921 $\pm$ 0.0022 \\
\cline{1-8}
\multirow{2}{*}{K-Nearest Neighbors} & NT &    0.534 $\pm$ 0.0057 &  0.894 $\pm$ 0.0022 &  0.556 $\pm$ 0.0075 &  0.943 $\pm$ 0.0020 &  0.436 $\pm$ 0.0640 &  0.845 $\pm$ 0.0026 \\
              & T &    0.557 $\pm$ 0.0056 &  0.884 $\pm$ 0.0023 &  0.569 $\pm$ 0.0074 &  0.940 $\pm$ 0.0020 &  0.446 $\pm$ 0.0693 &  0.835 $\pm$ 0.0027 \\
\cline{1-8}
\multirow{2}{*}{Random Forest} & NT &    0.590 $\pm$ 0.0080 &  0.900 $\pm$ 0.0027 &  0.598 $\pm$ 0.0094 &  0.947 $\pm$ 0.0017 &  0.618 $\pm$ 0.0960 &  0.938 $\pm$ 0.0019 \\
              & T &    0.474 $\pm$ 0.0007 &  0.901 $\pm$ 0.0026 &  0.485 $\pm$ 0.0005 &  0.942 $\pm$ 0.0019 &  0.318 $\pm$ 0.0435 &  0.840 $\pm$ 0.0033 \\
\cline{1-8}
\multirow{2}{*}{Gaussian Naive Bayes} & NT &    0.119 $\pm$ 0.0037 &  0.124 $\pm$ 0.0036 &  0.206 $\pm$ 0.0051 &  0.214 $\pm$ 0.0056 &  0.361 $\pm$ 0.1103 &  0.181 $\pm$ 0.0041 \\
              & T &    0.551 $\pm$ 0.0073 &  0.728 $\pm$ 0.0108 &  0.485 $\pm$ 0.0005 &  0.942 $\pm$ 0.0020 &  0.380 $\pm$ 0.0055 &  0.710 $\pm$ 0.0140 \\
\cline{1-8}
\multirow{2}{*}{Multinomial Naive Bayes} & NT &    0.569 $\pm$ 0.0062 &  0.762 $\pm$ 0.0061 &  0.463 $\pm$ 0.0083 &  0.655 $\pm$ 0.0109 &  0.421 $\pm$ 0.0047 &  0.793 $\pm$ 0.0059 \\
              & T &    0.572 $\pm$ 0.0063 &  0.771 $\pm$ 0.0060 &  0.463 $\pm$ 0.0083 &  0.655 $\pm$ 0.0109 &  0.421 $\pm$ 0.0048 &  0.792 $\pm$ 0.0060 \\
\cline{1-8}
\multirow{2}{*}{XGB} & NT &    0.569 $\pm$ 0.0069 & 0.906 $\pm$ 0.0026 &  0.576 $\pm$ 0.0097 &  0.947 $\pm$ 0.0019 &  0.912 $\pm$ 0.0144 &  0.941 $\pm$ 0.0023 \\
              & T       &    0.660 $\pm$ 0.0065 &  0.898 $\pm$ 0.0026 &  0.635 $\pm$ 0.0089 &  0.949 $\pm$ 0.0018 &  0.624 $\pm$ 0.0986 &  0.941 $\pm$ 0.0022 \\
\cline{1-8}
\multirow{2}{*}{Stochastic Gradient Descent} & NT &    0.502 $\pm$ 0.0128 &  0.901 $\pm$ 0.0026 &  0.594 $\pm$ 0.0115 &  0.945 $\pm$ 0.0018 &  0.746 $\pm$ 0.1193 &  0.916 $\pm$ 0.0032 \\
              & T &    0.591 $\pm$ 0.0383 &  0.834 $\pm$ 0.0699 &  0.595 $\pm$ 0.0147 &  0.946 $\pm$ 0.0017 &  0.746 $\pm$ 0.1258 &  0.916 $\pm$ 0.0033 \\
\cline{1-8}
\multirow{2}{*}{Gradient Boosting} & NT &    0.477 $\pm$ 0.0020 &  0.901 $\pm$ 0.0024 &  0.546 $\pm$ 0.0082 &  0.945 $\pm$ 0.0019 &  0.901 $\pm$ 0.0301 &  0.939 $\pm$ 0.0023 \\
              & T &    0.516 $\pm$ 0.0102 &  0.701 $\pm$ 0.0190 &  0.485 $\pm$ 0.0005 &  0.942 $\pm$ 0.0021 &  0.430 $\pm$ 0.0183 &  0.737 $\pm$ 0.0288 \\
\bottomrule
\end{tabular}
\vspace{4pt}
\caption{\car classification performance on \mimic.}
\label{mimic_classification_results}
\end{table*}